\documentclass{article}

\usepackage[final,nonatbib]{nips_2018}

\usepackage[utf8]{inputenc} 
\usepackage[T1]{fontenc}    
\usepackage{hyperref}       
\usepackage{url}            
\usepackage{booktabs}       
\usepackage{amsfonts}       
\usepackage{nicefrac}       
\usepackage{microtype}      
\usepackage{times}
\usepackage{epsfig}
\usepackage{graphicx}
\usepackage{amsmath}
\usepackage{url}
\usepackage{color}
\usepackage{subcaption}
\title{Revisiting the Importance of \\ Individual Units in CNNs via Ablation}

\author{
  Bolei Zhou \\
  MIT \\
  \And
  Yiyou Sun \\
  Harvard \\
  \And
  David Bau \\
  MIT \\
  \And
  Antonio Torralba \\
  MIT \\
}

\begin{document}

\maketitle

\begin{abstract}
We revisit the importance of the individual units in Convolutional Neural Networks (CNNs) for visual recognition. By conducting unit ablation experiments on CNNs trained on large scale image datasets, we demonstrate that, though ablating any individual unit does not hurt overall classification accuracy, it does lead to significant damage on the accuracy of specific classes. This result shows that an individual unit is specialized to encode information relevant to a subset of classes. We compute the correlation between the accuracy drop under unit ablation and various attributes of an individual unit such as class selectivity and weight L1 norm. We confirm that unit attributes such as class selectivity are a poor predictor for impact on overall accuracy as found previously in recent work \cite{morcos2018importance}. However, our results show that class selectivity along with other attributes are good predictors of the importance of one unit to individual classes. We evaluate the impact of random rotation, batch normalization, and dropout to the importance of units to specific classes. Our results show that units with high selectivity play an important role in network classification power at the individual class level. Understanding and interpreting the behavior of these units is necessary and meaningful.
\end{abstract}

\section{Introduction}

Attempts to understand the internal mechanisms of deep neural networks have frequently boiled down to analyzing the role of individual units. Previous work has visualized the units of deep convolutional networks by sampling image patches that maximize activation of each units \cite{zeiler2014visualizing,zhou2014object,girshick2016region} or by generating images that maximize each unit activation \cite{nguyen2016synthesizing}. Such visualizations show that individual units act as visual concept detectors.  Units at lower layers detect concrete patterns such as textures or shapes while units at high layers detect more semantically meaningful concepts such as dog heads or bicycle wheels. The interpretability of individual units has been quantified \cite{netdissect2017} by measuring the degree of alignment between their activations and human annotated concepts. However, though the units are shown to become more selective and more semantically meaningful in higher layers, how they contribute to the final prediction has not been analyzed in detail.

\begin{figure}[htpb]
\begin{center}
\includegraphics[width=1\linewidth]{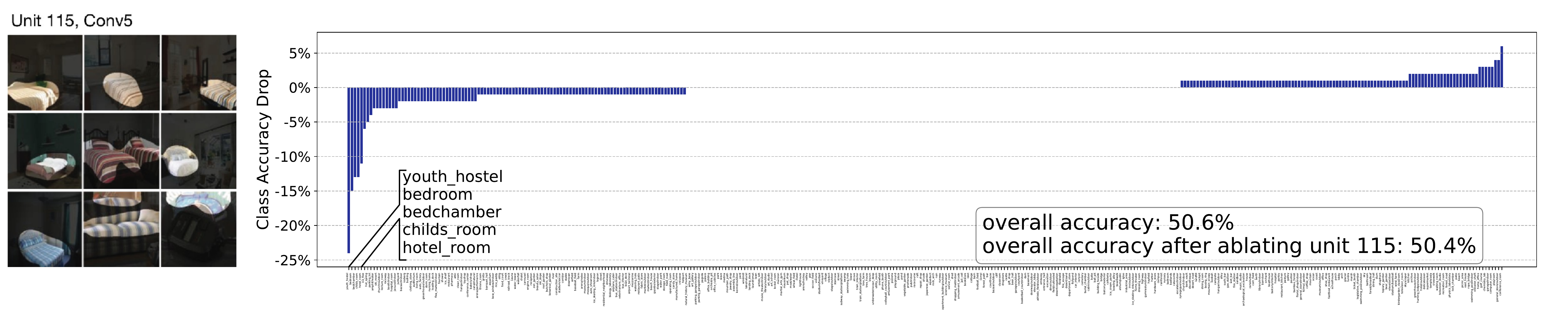}
\includegraphics[width=1\linewidth]{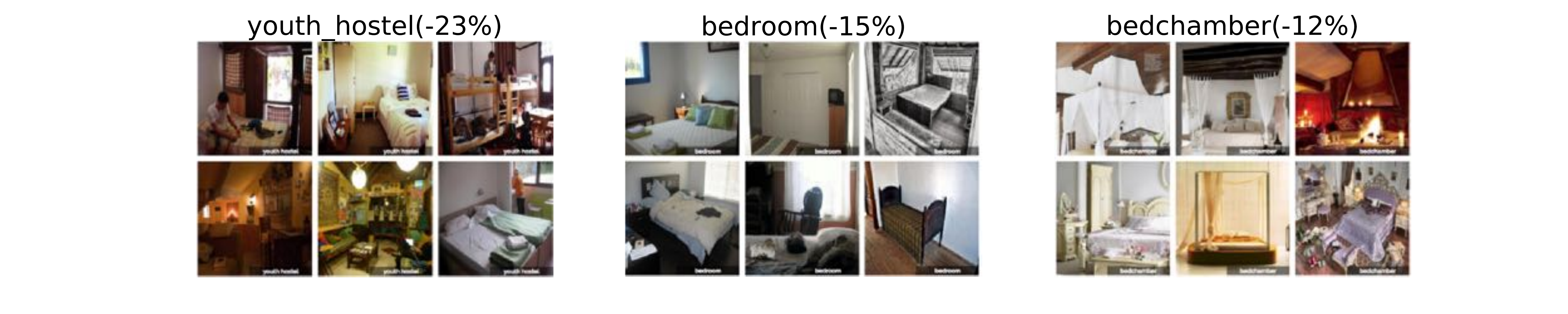}
\end{center}
\vspace{-4mm}
\caption{Ablating a unit which is selective to bed in images greatly damages the recognition of bed-relevant classes. We show 9 top activated images segmented by the feature activation on unit 115 from conv5 in AlexNet trained for place recognition. Ablating this unit from the network leads to a very small effect on the overall performance of the network (when classifying 365 place classes, it goes from 50.6\% accuracy to 50.4\% accuracy after ablating the unit.) Despite of the small drop (0.2\%) in overall accuracy, the drop in class accuracy is not uniform across all classes. As shown in the plot of sorted drop in class accuracy, the drop is significant ($>10\%$) in some classes such as \textit{youth hostel} and \textit{bedroom} for which detecting beds is important. Image samples from the top 3 damaged classes are shown below. }
\label{teaser}
\vspace{-3mm}
\end{figure}

A very interesting recent paper \cite{morcos2018importance} has used unit ablation experiments to quantify the importance of the units to the network output. The paper has shown that class selectivity is a poor predictor of a unit's importance towards overall accuracy of the classification task. These results suggest that `highly class selective units may be harmful to the network performance' implying that `understanding neural networks based on analyzing highly selective single units or finding optimal inputs for single units, may be misleading.' That conclusion seems to go against previous work on single unit visualization, opening a debate on the importance of understanding individual units in CNNs. In this paper we will show that both results are compatible. We have conducted a detailed analysis on the effect of ablation, showing that a different story is revealed when comparing the importance of highly selective units to their contribution to overall accuracy versus specific class accuracy. 

For example, Figure \ref{teaser} shows how ablating one unit has a severe impact on the accuracy in specific classes while showing only minimal effects on overall accuracy.  For instance, if we remove the unit which is selective to ‘beds’ the overall performance does not drop much. However the accuracy for recognizing the \textit{youth hostel} class drops by 23\% while the one for \textbf{bedroom} class drops by 15\%. That is a huge drop in performance for a single class given that we are only removing one unit. As there are 365 categories, doing badly on a single category (or a handful) might have almost no impact on the overall performance, but the classifier will perform really bad on some classes. 


The current work shows that units with high selectivity in a network are crucial for recognizing specific classes. We confirm that when ablating an individual unit, the damage to generalization accuracy averaged over all classes is small: this is consistent with the findings in \cite{morcos2018importance}. But we further find that ablating a single unit tends to cause significant damage to generalization accuracy on a subset of classes. We show that the class selectivity as well as other attributes are a reasonably good predictors for measuring the importance of the individual units to specific classification accuracy, and that understanding and interpreting the behaviors of individual units is meaningful.

\subsection{Related Work}

Investigation of deep network internals can be summarized at three major levels of detail as follows.

Whole networks have been studied to characterize their learning ability.  It has been demonstrated \cite{zhang2017understanding} that the same classification networks that successfully generalize meaningful labels are also able to memorize data with meaningless random labels.  Conversely, it has been observed \cite{szegedy2013intriguing} that networks are susceptible to being fooled on tiny perturbations of test inputs. These observations have challenged traditional assumptions about what a network learns, and why. Our current work contributes to the effort to characterize the mechanisms deep networks use to achieve generalization.

Networks have also have been studied at the granularity of individual layers.  It has been found that the representations produced by a layer summarize the lessons learned in training in a meaningful way \cite{yosinski2014transferable}: low-level layers can be used to summarize simpler patterns, and higher-level layers encode more abstract representations. Layers trained on one problem can be reused to transfer knowledge to other problems \cite{razavian2014cnn}, with a small amount of additional training. This fact has profound practical importance, reducing the need to build separate large training sets for every problem.

Finally, the role of individual network units has been studied.  It has been found that individual units are sensitive to specific concepts that can be visualized by generating or sampling inputs that maximize unit activations \cite{zhou2014object, nguyen2016synthesizing}, and that unit directions match meaningful concepts more closely than other random directions in representation space \cite{netdissect2017}.  However, whether meaningful units are helpful or harmful to a network's ability to generalize has been questioned: \cite{morcos2018importance} found that units that are selective to one class do not appear to damage overall classification performance more than other units when removed from a network. Our current work is a further examination of the impact of individual units on generalization accuracy: we ask, what is the role of a single unit in contributing to classification performance, and what characteristics of units are predictive of their contribution?



\section{Measuring the Importance of Units for Classification}

Given the human level visual recognition performed by deep neural network, without doubt each individual unit in the network plays a role in recognizing the image. However the importance of individual units on overall classification accuracy is not as well analyzed or quantified. In this section we measure the importance of units for classification through unit ablation and then characterize the relationship to several other relevant attributes.

\subsection{Classification Accuracy Drop from Unit Ablation}

To quantify the importance of an individual unit for the network's classification accuracy, we conduct unit ablation experiments with a similar setup as that used in \cite{morcos2018importance}. We ablate one unit then compute the resulting classification accuracy drop.  This provides a way to quantify the importance of unit in the classification. Note that for all experiments we measure \textit{generalization accuracy}, which is the classification accuracy on a held-out validation set, rather than accuracy on the training set.

We ablate a unit as follows: For each unit, we set its weight and bias as 0 so that it will not contribute to the prediction for any input image. We pass the images from the validation set into the network with one unit ablated at a time, then compute the generalization accuracy drop. 

In our analysis, we measure two types of accuracy drop: the first one is the \textbf{overall accuracy drop} which is the total percentage of images misclassified due to ablating the unit. The second one is \textbf{class accuracy drop} which measures the difference in accuracy in correctly classifying images of each individual class due to ablating the unit. If the number of testing samples per class is the same, which is the case for both ImageNet \cite{russakovsky2015imagenet} and Places \cite{zhou2016places}, then the overall accuracy drop is simply average of the class accuracy drops. Note that the class accuracy drop is a vector with dimension equal to the number of classes. We can summarize this vector by its maximum value, which is the \textbf{max class accuracy drop}.  The max class accuracy drop measures the importance of unit to predictions of instances of an individual class. We can further sort units in a layer by their max class accuracy drop to have \textbf{max class accuracy drop curve}, as a measure for the degree of class-specific information carried by individual units. We will show later in Fig.\ref{figure_selectivity_sample} and Fig.\ref{figure_generalization}, the more flat the curve is, the less class-specific information carried by individual units, therefore units will be less in single meaningful directions and less interpretable in a layer. 

In later experiments, we will build a model to predict the overall accuracy drop and max class accuracy drop based on other observable characteristics of a unit. We will show that although ablating an individual unit does not tend to cause a large drop in overall accuracy, it does lead to significant loss in accuracy for a subset of classes.  This suggests that individual units align with single directions in representation space that carry information that is specialized to a subset of classes.

\subsection{Characterizing the Individual Units with Attributes}

To further understand the reasons for a unit's importance to class accuracy and overall generalization accuracy, we study several other attributes to characterize each individual unit. We would like to understand how these attributes of units may be correlated with or predictive of the unit's importance in classification.

In various different contexts in the literature, several metrics have been used to quantify the properties of individual units: for example class selectivity \cite{morcos2018importance}, unit correlation \cite{li2015convergent}, the alignment between unit activation and visual concepts used in network dissection \cite{netdissect2017}, or L1 norm of the filter weight which is commonly used in network compression \cite{li2016pruning,han2015learning}. We summarize each attribute as follows:

\textbf{L1 Norm}. This number is simply the L1 norm of the learned weights of the unit. Despite of the simplicity of the metric, it has been shown to be an effective metric for pruning units in network compression \cite{han2016dsd}. The L1 norm of the $i$th unit is
\begin{align}
\mathrm{norm}_1(i) = ||w_i||_1 = \sum_{j} |(w_i)_j|
\end{align}
\textbf{Class Correlation}. As shown in \cite{li2015convergent}, the statistical correlation of unit activations reflects how related two units' representations of images are. So that we can use the correlation between the activation of unit $i$ and the predicted probability for class $k$ as the amount of information carried by the unit. The class correlation is computed as 
\begin{align}
\mathrm{corr}(i, k) = \frac{E[(x_i - \bar{x}_i)(p_k - \bar{p}_k)]}{\sigma_{x_i} \sigma_{p_k}}
\end{align}
where $x_i$ denotes the $i$th unit's activations, $p_k$ denotes the $k$th class's predicted probability for the same set of input images, and $\bar{x}_i$ and $\bar{p}_k$ denote mean values while $\sigma_{x_i}$ and $\sigma_{p_k}$ denote the standard deviation. Note that for each unit, the class correlation is a vector of dimension equal to the number of classes. We can take the maximum value of the vector as the scalar class correlation.

\textbf{Class Selectivity}. The class selectivity is a metric for measuring the single direction of units used in \cite{morcos2018importance}. It is calculated as follows,
\begin{align}
\mathrm{select}(i, k) = \frac{\bar{x}_{i}^{k} - \bar{x}_{i}^{-k}}{\bar{x}_{i}^{k}+\bar{x}_{i}^{-k}}
\end{align}
where $\bar{x}_{i}^{k}$ denotes the mean activation value of unit $i$ when the input image belong to class $k$ and $\bar{x}_{i}^{-k}$ denotes the mean activation value of unit $i$ averaged across all other classes (non $k$th classes). Refer to \cite{morcos2018importance} for detailed description of the metric. The class selectivity is also a vector of dimension equal to the number of classes. We can take the maximum value of the vector as the scalar class selectivity as used in \cite{morcos2018importance}. This metric varies from 0 to 1, with 1 indicating the unit active for inputs of a single class and 0 indicating the unit active identically for all classes.

\textbf{Concept Alignment}. In \cite{netdissect2017}, the interpretability of units is measured as the IoU (Intersection over Union) score, the degree of alignment between unit activation and the ground-truth semantic mask containing 1198 visual concepts. For each unit $i$, we can compute the IoU for each one of 1198 concepts used in \cite{netdissect2017} as the Concept IoU attribute of the unit. Taking the maximum of the IoU and its associated label, we get a scalar concept IoU of that unit as used in \cite{netdissect2017}.

\textbf{Unit Visualization}. Visualization of units can be also considered as a qualitative attribute that can reveal which image patterns are represented by the unit. To visualize a unit we generate the top activated images following \cite{zhou2014object,netdissect2017,girshick2014rich} by showing the images in the validation set that most activate the unit. These images are further segmented by applying a binarized activation map to highlight the most activated image regions. Though there are other way to visualize unit such as back-propagation \cite{simonyan2013deep,olahfeature} or image synthesis \cite{nguyen2016synthesizing}, we find this instance-based visualization intuitive.

We characterize the individual units using the attributes above. There are other statistical properties that can be used to characterize unit activation, such as the sparsity, mean, and standard deviation of the activation distributions.
In Sec.\ref{sec:relation} we analyze the relation between the importance of units in classification and the attributes described above. We will further show that some of the attributes are a good predictors of unit importance.

\section{Experiments}

\subsection{Models and Datasets}

For the following experiments, we use AlexNet \cite{krizhevsky2012imagenet} and ResNet18 \cite{he2016deep} trained from scratch on ImageNet \cite{russakovsky2015imagenet} and Places \cite{zhou2016places} as the testing models.
ImageNet has 1.2 million training images from 1,000 object classes while Places has 1.6 million training images from 365 scene classes. For the validation set used to compute the accuracy drop from unit ablation, there are 50 images per class in ImageNet and 200 images per class in Places.

\subsection{Influence of Unit Ablation to Accuracy Drop}
\label{sec:ablation}
\begin{figure}
\begin{center}
\vspace{-5mm}
\includegraphics[width=1\linewidth]{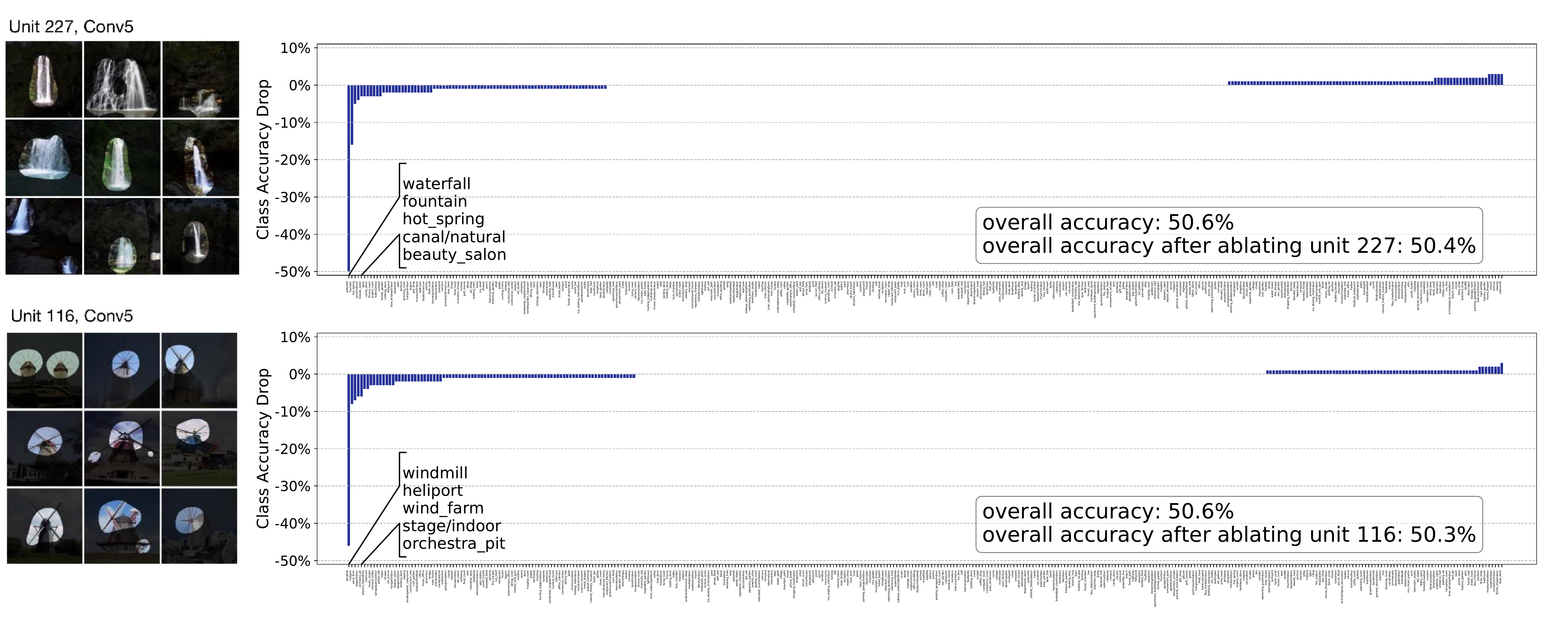}
\includegraphics[width=1\linewidth]{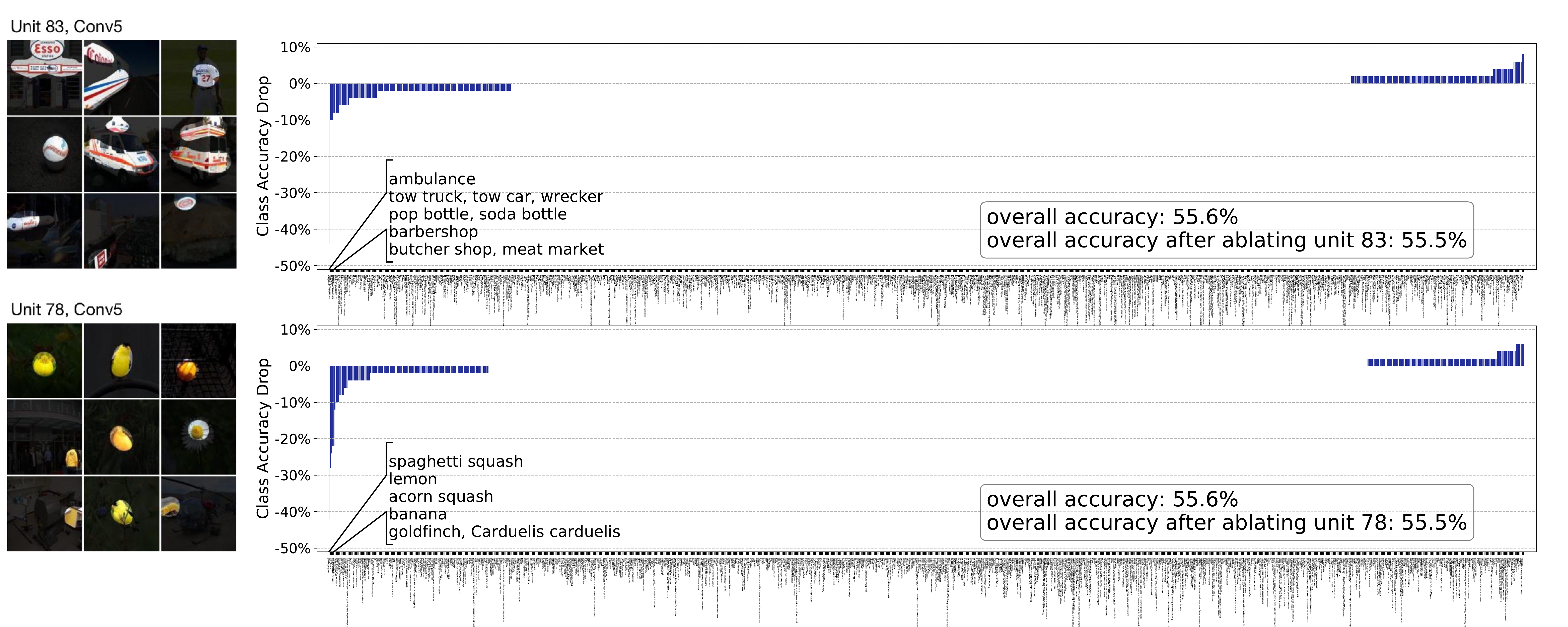}
\includegraphics[width=1\linewidth]{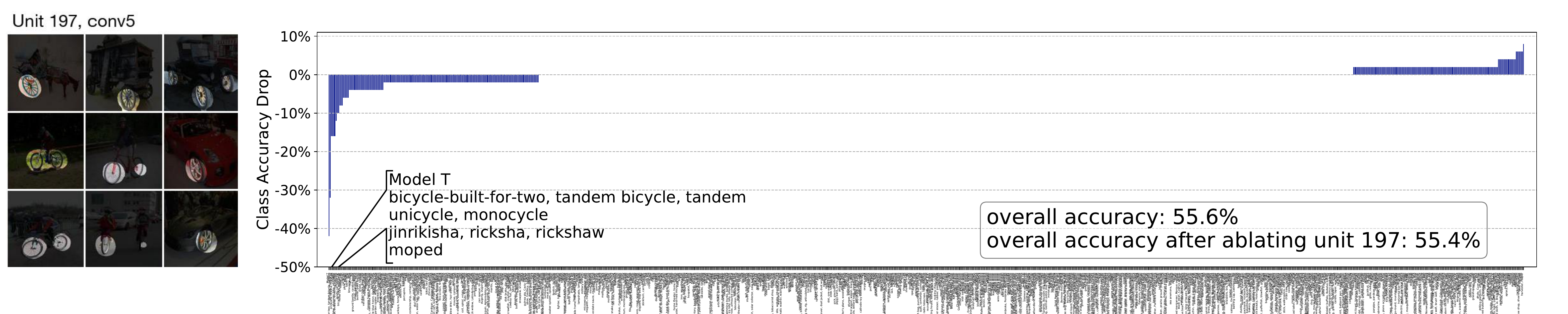}
\includegraphics[width=1\linewidth]{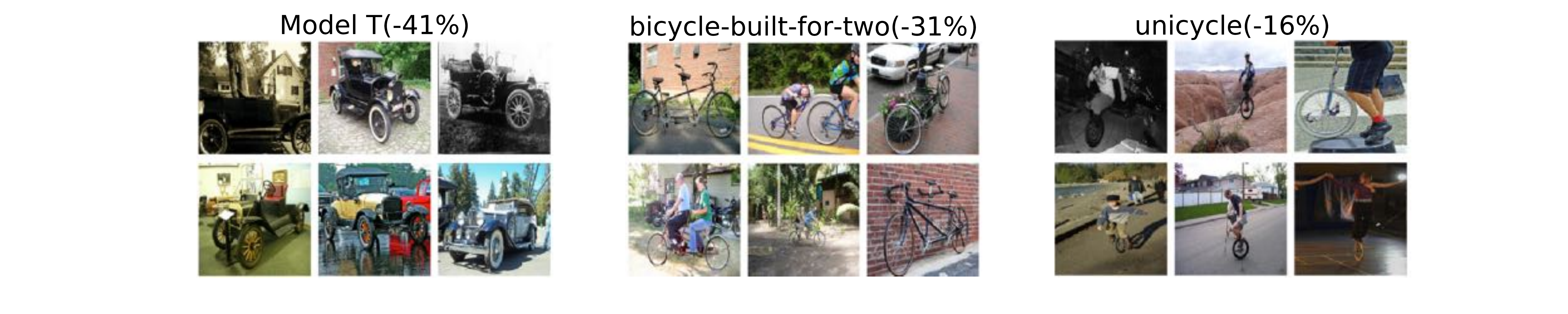}
\end{center}
\vspace{-5mm}
\caption{Ablated units from AlexNet-Places (top two units) and AlexNet-ImageNet (bottom three units) respectively. For the last unit in AlexNet-ImageNet which is selective to wheels, we also show the image samples from the top three damaged classes in which wheels are important parts to represent the classes. For each unit ablated, we show the unit visualization, the sorted class accuracy drops, and the overall accuracy before and after unit ablation. We can see that ablating one unit brings a significant class accuracy drop to some specific classes, compared to the trivial drop in overall accuracy. The most damaged classes match the unit visualization very well. A few classes also benefit slightly from the unit ablation, though they are not semantically related. We check the categories which benefit from the unit ablation and they are not meaningful so that we don't show them in detail. When averaging class accuracy drop to obtain the overall accuracy drop, the overall drop is small.}
\label{figure_ablation_qualitative_both}
\vspace{-4mm}
\end{figure}

For each ablated unit, we compute the resulting overall accuracy drop and class accuracy drop. Some examples of unit ablations are shown in Fig.\ref{figure_ablation_qualitative_both}. In this figure, we sort all the classes by class accuracy drop. We can see that some classes are damaged significantly when ablating a single unit. For example, ablating a unit which is sensitive to waterfalls (as seen in the visualization) hurts the categories \textit{waterfall}, \textit{fountain}, \textit{hot spring}.  Specifically, the category of \textit{waterfall} suffers a huge 50\% drop in class accuracy.  In the third example, ablating a unit that is sensitive to trade signs on a white background impacts the categories of \textit{ambulance}, \textit{tow truck}, and \textit{pop bottle} significantly.

\begin{figure}
\begin{center}
\vspace{-1mm}
\includegraphics[trim={4cm 0 4cm 0},width=0.9\linewidth]{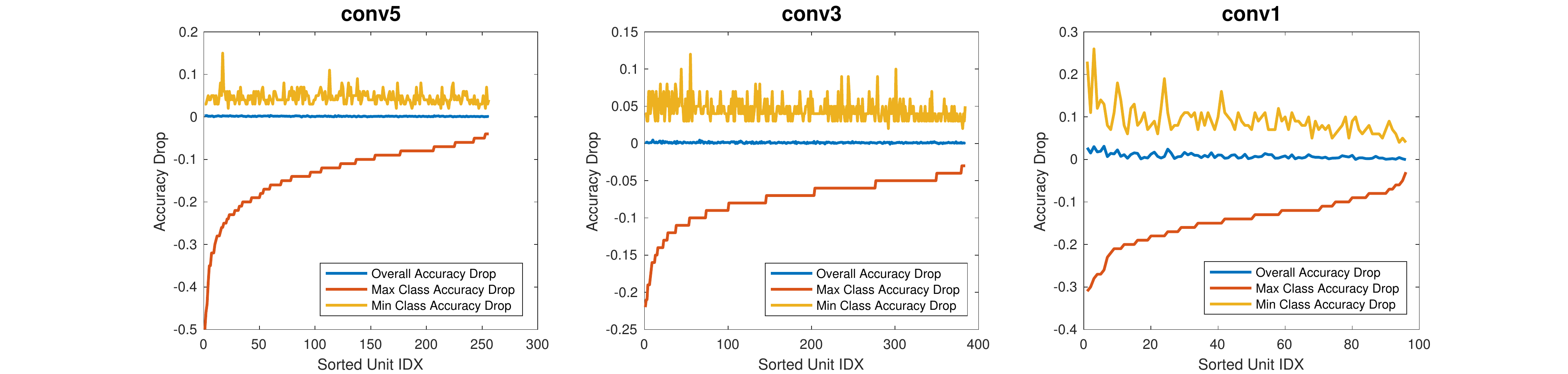}
\end{center}
\vspace{-4mm}
\caption{Units at each one of the three layers from AlexNet-Places are sorted by their max class accuracy drops. For each unit we also plot the min class accuracy drop and the overall accuracy drop. Ablating units has different effect on overall accuracy and class accuracy.}
\label{figure_sortablation}
\vspace{-3mm}
\end{figure}

For all the units at one layer, we take the max class accuracy drop as well as the min class accuracy drop for each unit then sort all the units by their max class accuracy. Note that the min class accuracy drop reveals a slight gain in accuracy for some classes when a unit is ablated. This provides an overview of each unit's importance to individual classes.  The result on the three convolutional layers of AlexNet-Places is plotted in Fig.~\ref{figure_sortablation}. The drop in overall accuracy (blue curve) is very small (below $1\%$), which is in agreement with results shown in \cite{morcos2018importance}.  Regardless of the characteristics of a unit, the effect of a single unit ablation on overall accuracy is marginal. However, observing max class accuracy drop (red curve) reveals a different picture. The drop in performance for individual classes can be very significant. For instance, in the conv5 layer, the max drop is $50\%$, while more than half of the units produce a drop above $10\%$ on individual classes if ablated. As shown previously in Fig.\ref{teaser} removing a `bed' selective unit strongly affects the recognition of bedrooms and hotel rooms. The overall impact on the classification accuracy for the other categories is negligible, but a network missing that unit will have a higher error rate when tested on bedrooms. These results show that individual units tend to specialize: each unit plays an important role in recognizing a subset of specific classes. Therefore, any analysis of unit importance should examine the effect on individual classes; an analysis based only on overall performance can miss important effects when manipulating a network.

\textbf{Greedy Unit Ablation}. To further explore how groups of units collectively contribute to predictions for a class, we conduct an experiment in which we ablate a series of units successively.  In this experiment, we follow a greedy approach, iteratively ablating additional units that maximally reduce the accuracy of a specific class.

\begin{figure}
\begin{center}
\includegraphics[width=0.6\linewidth]{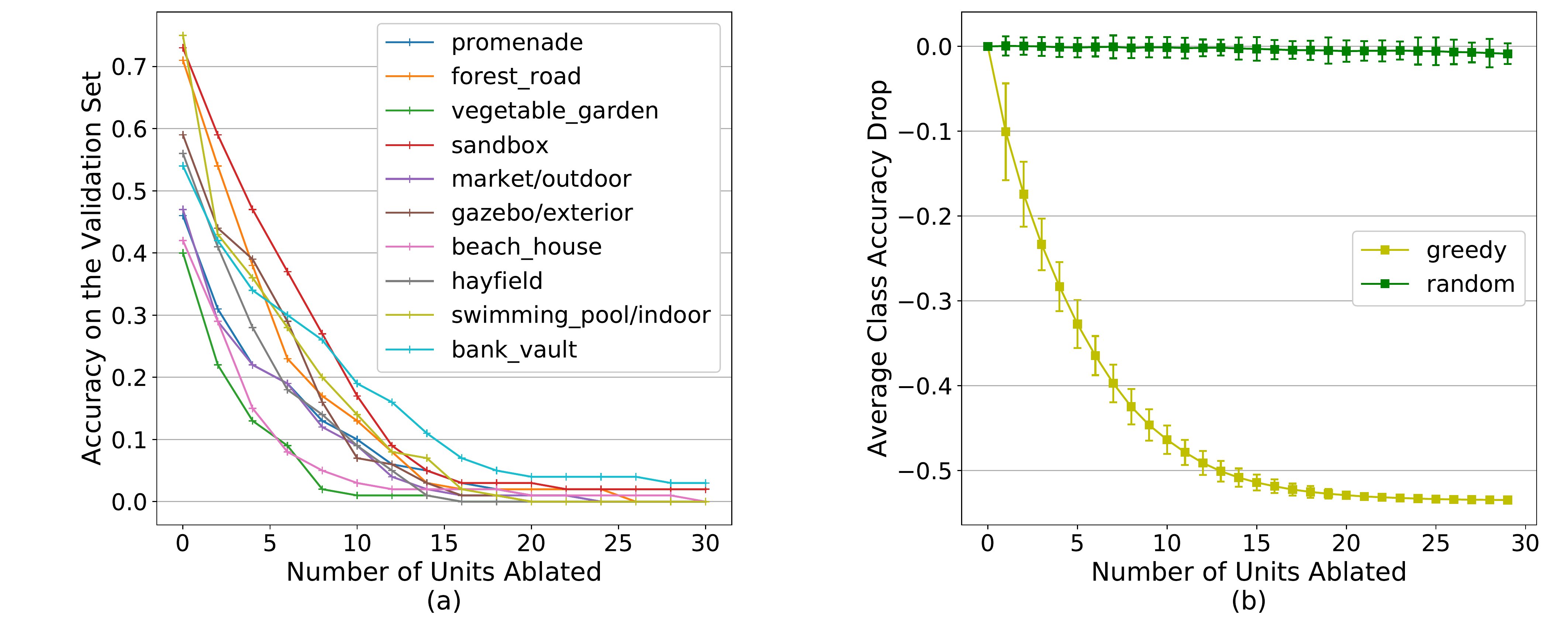}
\end{center}
\vspace{-4mm}
\caption{a) The class accuracy over the number of units ablated for a random selection of classes from Places. b) The average max class accuracy drop over the number of units ablated in greedy unit ablation. }
\label{result_greedy_removal}
\vspace{-4mm}
\end{figure}

The result of greedy unit ablation on conv5 layer of AlexNet-Places is shown in Fig.\ref{result_greedy_removal}. Fig.\ref{result_greedy_removal}a shows that class accuracy drops sharply when the number of units ablated increases, with accuracy for a specific class dropping to very low levels after 10-15 units are ablated. In Fig.\ref{result_greedy_removal}b, we further plot class accuracy drop over the number of units ablated averaged across all the categories (for each class we conduct a separate series of greedy unit ablations).  This is compared with the effect of ablations of a randomly chosen sequence of units as a baseline comparison.

We find that ablating top 5 most informative units for a class brings about 28\% class accuracy drop, and top 10 most informative units units for more than 40\% class accuracy drop. On the contrary there is negligible change in class accuracy if a random set of units is ablated. 

\subsection{Relation between the Accuracy Drop and other Attributes}\label{sec:relation}

What attributes does a unit have that make it important to class-specific and overall generalization accuracy?  For insight, we analyze the relation between a set of unit attributes and the ablation accuracy drops. We examine unit class correlation and class selectivity in terms of the class with maximum value; for concept IoU we examine the concept with maximum IoU value.

\begin{figure}
\begin{center}
\vspace{-4mm}
\includegraphics[trim={4cm 0 4cm 0},clip,width=1\linewidth]{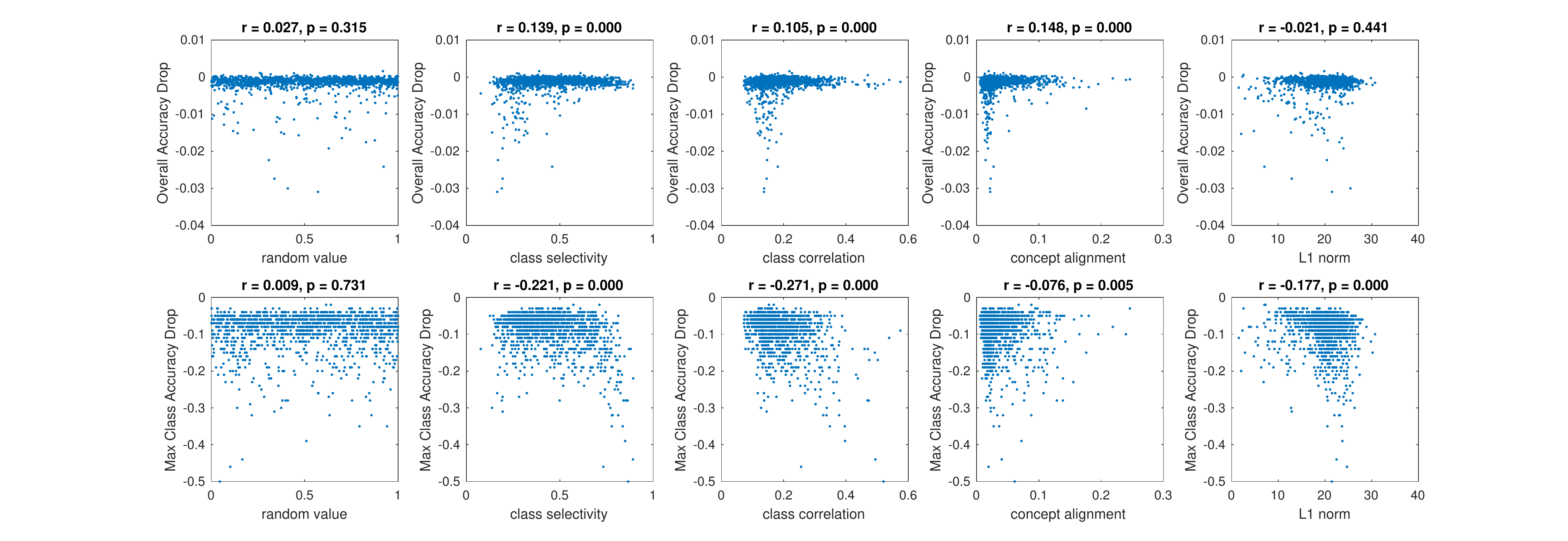}
\end{center}
\vspace{-6mm}
\caption{Overall accuracy drop (the upper row) and max accuracy drop (the lower row) over the unit attributes for all the units in AlexNet-Places. $r$ is the Spearman's rank-order correlation and $p$ is the $p$-value. We also include a random value baseline in the first column.}
\label{figure_correlation_attributes}
\vspace{-1mm}
\end{figure}

The results on AlexNet-Places365 are shown in Fig.\ref{figure_correlation_attributes}. We can see that for overall accuracy drop (the upper row), there are positive correlations for class selectivity, class correlation and concept alignment.  That is, as these measures of alignment between a unit and a single concept decrease, the unit tends to cause a larger drop in overall accuracy when ablated. This finding is consistent with the result in \cite{morcos2018importance}, where it was found that highly class selective units cause less damage to overall accuracy when ablated. As argued in \cite{morcos2018importance} this result questions the necessity of studying the units with high single direction selectivity as suggested in \cite{zeiler2014visualizing,zhou2014object}. Unit L1 stands out as the one metric that correlates with overall accuracy drop.

On the other hand, examining the correlation between the unit attributes and max class accuracy drop shown at the lower row in Fig.\ref{figure_correlation_attributes} shows a different story. There is a clear negative correlation between the max class accuracy drop and each of the measured attributes. Class selectivity and class correlation show particularly strong negative correlations: the more closely aligned a unit is with a class, the larger the max class accuracy drop. Alignment of a unit with a visual concept is also correlated with max class accuracy drop, but not as strongly, indicating that concept-aligned units spread out their contribution to many classes.  Interestingly we also find that L1 norm is a good metric for unit importance in class accuracy as well.  The utility of L1 to predict both overall and max class accuracy impact may be a reason for the success of the L1 norm in network compression.

\textbf{Predicting the class one unit contributes most using the attributes}. From our previous analysis, we show that an individual unit is specialized to represent one or a subset of classes while the attributes are correlated with the max class accuracy drop. Here we further use the unit attributes to predict the class one individual unit contributes most, to see how predictable different metrics are. 

We prepare five instances of AlexNet-Places trained with different random initialization. We use the attributes of units from conv5 layer of three network instances for training; then we evaluate the model on the units from conv5 layer of the other two network instances. 

We test the use of three attributes Class Correlation, Class Selectivity, and Concept Alignment, as alternative features for training a simple logistic regression classifier. Note that we use the vector of each attributes rather than taking the maximum value as the features. The ground-truth label for each unit is the index of the class which has the largest accuracy drop when that unit is ablated.

\begin{figure}[h]
\vspace{-2mm}
\begin{center}
\includegraphics[width=1\linewidth]{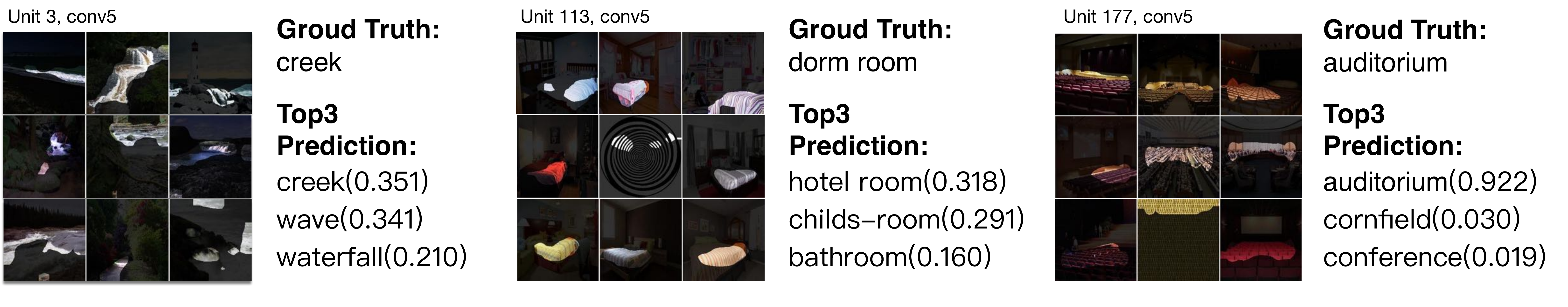}
\end{center}
\vspace{-3mm}
\caption{Examples of predicting the class one unit contributes most. The ground-truth class is from the class with max class accuracy drop when the unit is ablated. Top 3 predictions are given by the classifier trained on Concept Alignment scores. }
\label{figure_predictioin}
\vspace{-3mm}
\end{figure}

Prediction accuracy using different attributes is shown in Table \ref{table:att_acc}. We can see that Concept Alignment outperforms the other two attributes. This number exceeds the correlation measurement in Sec.~\ref{sec:relation} because the model can adjust its sensitivity to aligned concepts according to their specific relevance to classes in the target problem.  The reasonably good prediction accuracy suggest that the units with both high class selectivity and concept interpretability play an important role in recognizing specific classes. Some prediction examples are shown in Fig.\ref{figure_predictioin}.

\begin{table}
\begin{center}
\begin{tabular}{l c c c c}
 \hline
 Attribute & Concept Alignment & Class Correlation & Class Selectivity & Ensemble \\
 \hline
 Accuracy & 38.67\% & 38.28\% & 38.08\% & 39.06\% \\
\hline
\end{tabular}
\caption{Accuracy for predicting the class one unit contributes most using different attributes. Evaluated on the units of two held-out network instances, the attributes are reasonably well predictors for the unit importance to class. Here ensemble takes the average of the predicted probabilities from the previous three models.}
\label{table:att_acc}
\end{center}
\vspace{-8mm}
\end{table}

\subsection{Unit Ablation Under Random Rotation}

We have shown that some individual units are important for the generalization accuracy of specific classes, which suggests the role of individual units is to specialize in information that is useful for only a subset of decisions.  To verify that this phenomenon is specific to unit directions and not merely a result of ablating random directions, we compare the measurements to a randomized baseline.

First, we obtain random directions in representation space by applying a random rotation on the learned units, similar to the random projection of units in \cite{netdissect2017}. The random rotation of units creates a new set of random directions (the same as the number of units). Applying the inverse rotation recovers the original representation, so classification accuracy is not affected. To ablate a random direction, we zero a coordinate in the rotated representation, erasing the selected random direction.

\begin{figure}
\begin{center}
\vspace{-5mm}
\includegraphics[width=1\linewidth]{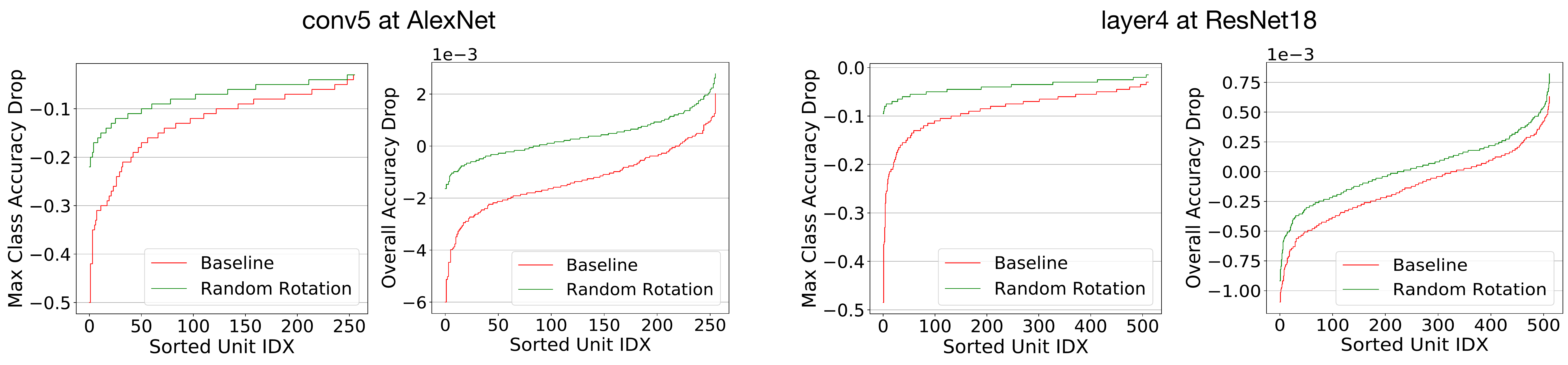}
\end{center}
\vspace{-4mm}
\caption{Comparison of the max class accuracy drop curves for random directions in representation space, on the conv5 layer of AlexNet and layer4 of ResNet18. The left image in each group: Sorted max class accuracy drop of unit ablations versus ablations of a basis of random directions in representation space. The right image in each group: a similar comparison for overall accuracy across all classes.  Note that the scales of the second and fourth graph are amplified by 1000 times to make the small gap visible.}
\label{figure_selectivity_sample}
\vspace{-3mm}
\end{figure}

The max class accuracy drop curves are shown in first and third graphs in Fig.~\ref{figure_selectivity_sample}. It can be seen that units are more specialized towards individual classes than random rotated directions in representation space, providing support to the hypothesis that generalization accuracy of individual classes does rely on specific units. The second and fourth graphs in Fig.~\ref{figure_selectivity_sample} shows a similar comparison with respect to the overall accuracy drop.  Although again we find individual unit directions appear more important than arbitrary directions in representation space, notice that the scale is different: the comparative importance of individual units over random directions for overall generalization performance is small.

These comparisons reveal that not only are unit directions more important for generalization than random directions in representation space, but that the contributions of most unit directions are systematically focused on a subset of classes rather than generically on all classes.

\subsection{Influence of Batch Normalization and Dropout on the Units}

\begin{figure}
\begin{center}
\includegraphics[width=1\linewidth]{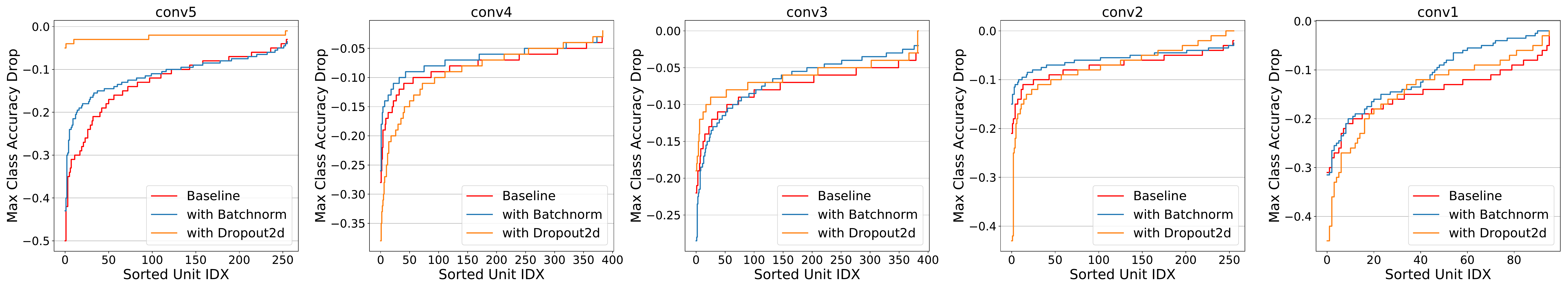}
\end{center}
\vspace{-4mm}
\caption{Comparison of the max class accuracy drop curves for the baseline, network with batch normalization, and network with dropout. Since Batch norm is on all the conv layers, it reduces the class-specific information carried by the individual units across all the layers. Dropout on conv5 greatly reduces the max class accuracy drop for all the units.}
\label{figure_generalization}
\vspace{-5mm}
\end{figure}

\begin{figure}[h]
\begin{center}
\vspace{-2mm}
\includegraphics[width=1\linewidth]{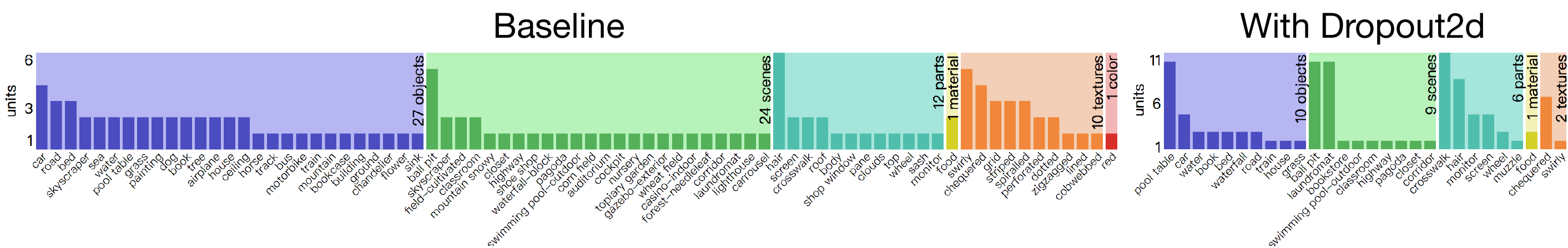}
\end{center}
\caption{Histogram of interpretable units identified by network dissection \cite{netdissect2017} on the conv5 layer of the baseline and the network with channel dropout on conv5. Dropout greatly reduces the number of interpretable units in a layer. }
\vspace{-5mm}
\label{figure_netdissect}
\end{figure}

Batch normalization \cite{ioffe2015batch} and dropout \cite{srivastava2014dropout} are effective training regularizers for deep neural networks. Batch normalization \cite{ioffe2015batch} speeds up the training process and enables the training of much deeper networks \cite{he2016deep}, while dropout randomly blackout features in training to prevent overfitting \cite{srivastava2014dropout}. However, the impact of these two on the representation is not as well understood \cite{zhang2017understanding,morcos2018importance}. 
Here we compare the unit ablation on representation trained with batch normalization and with dropout. The baseline is the naive AlexNet-Places. To evaluate the batch norm, We add batch normalization on each of the 5 conv layers on the baseline. The network with batch norm achieves similar generalization accuracy with the original baseline. To evaluate the dropout, we add channel dropout on conv5 layer only (in each training batch 50\% of channels/units are randomly dropped out; thus in experiment, dropout is applied to conv5, fc6, and fc7, wheras in baseline only fc6 and fc7 have dropout). The network with channel dropout achieved 2\% lower accuracy. The max class accuracy drop curves are shown in Fig.\ref{figure_generalization}. Compared with the baseline, we can see that batch normalization and dropout reduce class information contained in each individual unit. Interestingly dropout greatly reduces max class accuracy drop for all the units, scattering class-specific information across all the units rather than being concentrated. That reduces the alignment of individual units at conv5 with single interpretable directions (see the network dissection result in Fig.\ref{figure_netdissect}). However dropout on conv5 layer increases max class accuracy drop at lower conv4 layer as compensation. Thus batch normalization and dropout, initially proposed to address the issue of training efficiency and overfitting, also affect class-specific information carried by the units and their interpretability. Designing a regularizer that improves both the classification performance and interpretability of individual units remains an open problem. 

\section{Conclusion}

By conducting a set of unit ablation experiments, we have refined our understanding of the importance of individual units on the generalization accuracy of CNNs trained for visual recognition. We have verified that removing one unit does not significantly damage overall network generalization accuracy, which is consistent with \cite{morcos2018importance}. However, we have also found that removing an individual unit causes significant damage to the accuracy of a subset of classes.  The finding that individual units specialize in subsets of the classification space indicates that the contribution of individual units towards the decisions of a CNN is both nontrivial and specific.  Understanding how individual units decompose a network's predictions remains an important area for further study.

\textbf{Acknowledgement} This work was partially funded by DARPA XAI program to A.T.;This work was also partly supported  by the National Science Foundation under Grants No. 1524817 to A.T.. B.Z is supported by a Facebook Fellowship.  

{\small
\bibliographystyle{ieee}
\bibliography{egbib}
}

\end{document}